# Comparative Experiments on Disambiguating Word Senses: An Illustration of the Role of Bias in Machine Learning

**Raymond J. Mooney**
Department of Computer Sciences
University of Texas
Austin, TX 78712-1188
mooney@cs.utexas.edu

## Abstract

This paper describes an experimental comparison of seven different learning algorithms on the problem of learning to disambiguate the meaning of a word from context. The algorithms tested include statistical, neural-network, decision-tree, rule-based, and case-based classification techniques. The specific problem tested involves disambiguating six senses of the word "line" using the words in the current and proceeding sentence as context. The statistical and neural-network methods perform the best on this particular problem and we discuss a potential reason for this observed difference. We also discuss the role of *bias* in machine learning and its importance in explaining performance differences observed on specific problems.

## Introduction

Recent research in empirical (corpus-based) natural language processing has explored a number of different methods for learning from data. Three general approaches are statistical, neural-network, and symbolic machine learning and numerous specific methods have been developed under each of these paradigms (Wermter, Riloff, & Scheler, 1996; Charniak, 1993; Reilly & Sharkey, 1992). An important question is whether some methods perform significantly better than others on particular types of problems. Unfortunately, there have been very few direct comparisons of alternative methods on identical test data.

A somewhat indirect comparison of applying stochastic context-free grammars (Periera & Shabes, 1992), a transformation-based method (Brill, 1993), and inductive logic programming (Zelle & Mooney, 1994) to parsing the ATIS (Airline Travel Information Service) corpus from the Penn Treebank (Marcus, Santorini, & Marcinkiewicz, 1993) indicates fairly similar performance for these three very different methods. Also, comparisons of Bayesian, information-retrieval, neural-network, and case-based methods on word-sense disambiguation have also demonstrated similar performance (Leacock, Towell, & Voorhees, 1993b; Lehman, 1994). However, in a comparison of neural-network and decision-tree methods on learning to generate the past tense of an English verb, decision trees performed significantly better (Ling & Marinov, 1993; Ling, 1994). Subsequent experiments on this problem have demonstrated that an inductive logic programming method produces even better results than decision trees (Mooney & Califf, 1995).

In this paper, we present direct comparisons of a fairly wide range of general learning algorithms on the problem of discriminating six senses of the word "line" from context, using data assembled by Leacock et al. (1993b). We compare a naive Bayesian classifier (Duda & Hart, 1973), a perceptron (Rosenblatt, 1962), a decision-tree learner (Quinlan, 1993), a $k$ nearest-neighbor classifier (Cover & Hart, 1967), logic-based DNF (disjunctive normal form) and CNF (conjunctive normal form) learners (Mooney, 1995) and a decision-list learner (Rivest, 1987). Tests on all methods used identical training and test sets, and ten separate random trials were run in order to measure average performance and allow statistical testing of the significance of any observed differences. On this particular task, we found that the Bayesian and perceptron methods perform significantly better than the remaining methods and discuss a potential reason for this observed difference. We also discuss the role of *bias* in machine learning and its importance in explaining the observed differences in the performance of alternative methods on specific problems.

## Background on Machine Learning and Bias

Research in machine learning over the last ten years has been particularly concerned with experimental comparisons and the relative performance of different classification methods (Shavlik & Di-etterich, 1990; Kulikowski & Weiss, 1991; Langley, 1996). In particular, the UCI Machine Learning Data Repository (Merz, Murphy, & Aha, 1996) was assembled to facilitate empirical comparisons. Experimental comparisons of different methods on various benchmark problems have generally found relatively small differences in predictive accuracy (Mooney, Shavlik, Towell, & Gove, 1989; Fisher & McKusick, 1989; Weiss & Kapouleas, 1989; Atlas, Cole, Conner, El-Sharkawi, Marks, Muthusamy, & Bernard, 1990; Dietterich, Hild, & Bakiri, 1990; Kulikowski & Weiss, 1991; Shavlik, Mooney, & Towell, 1991; Holte, 1993). However, on specific problems, certain methods can demonstrate a significant advantage. For example, on the problem of detecting promoter sequences in DNA (which indicate the start of a new gene), neural-network and similar methods perform significantly better than symbolic induction methods (Towell, Shavlik, & Noordewier, 1990; Baffes & Mooney, 1993). On the other hand, as mentioned in the introduction, symbolic induction methods perform significantly better than neural-networks on the problem of learning to generate the past tense of an English verb (Ling & Marinov, 1993; Ling, 1994; Mooney & Califf, 1995).

It is generally agreed that the philosophical *problem of induction* (Hume, 1748) means that no inductive algorithm is universally better than any other. It can be proven that when averaged over a uniform distribution of all possible classification problems, the generalization performance (predictive accuracy on unseen examples) of any inductive algorithm is zero. This has been called the "Conservation Law for Generalization Performance" (Schaffer, 1994) or a "no free lunch" theorem (Wolpert, 1992). However, averaging over a uniform distribution of all possible functions is effectively equivalent to assuming a "random universe" in which the past is not predictive of the future. If all problems are not equally likely, the *expected* generalization performance over a distribution of real-world problems can of course be positive (Rao, Gordon, & Spears, 1995).

In machine learning, *bias* refers to "any basis for choosing one generalization over another, other than strict consistency with the instances" (Mitchell, 1980). Decision-tree methods have a bias for simple decision trees, rule induction methods have a bias for simple DNF expressions, neural-network methods have a bias for linear threshold functions, [1] and naive Bayes has a bias for functions which respect conditional independence of features. The more the bias of a certain learning algorithm fits the characteristics of a particular problem, the better it will perform on that problem. Most learning algorithms have some sort of "Occam's razor" bias in which hypotheses that can be represented with fewer bits in some particular representation language are preferred (Blumer, Ehrenfeucht, Haussler, & Warmuth, 1987). However, the compactness with which different representation languages (e.g. decision trees, DNF, linear threshold networks) can represent particular functions can vary dramatically (e.g. see Pagallo and Haussler (1990)). Therefore, different biases can perform better or worse on specific problems. One of the main goals of machine learning is to find biases that perform well on the distribution of problems actually found in the real world.

As an example, consider the advantage neural-networks have on the promoter recognition problem mentioned earlier. There are several potential sites where hydrogen bonds can form between the DNA and a protein and if enough of these bonds form, promoter activity can occur. This is represented most compactly using an M-of-N classification function which returns true if any subset of size M of N specified features are present in an example (Fisher & McKusick, 1989; Murphy & Pazzani, 1991; Baffes & Mooney, 1993). A single linear threshold unit can easily represent such functions, whereas a DNF expression requires "N choose M" terms to represent them. Therefore, the difference in their ability to compactly represent such functions explains the observed performance difference between rule induction and neural-networks on this problem. [2]

Of course picking the right bias or learning algorithm for a particular task is a difficult problem. A simple approach is to automate the selection of a method using internal cross-validation (Schaffer, 1993). Another approach is to use meta-learning to learn a set of rules (or other classifier) that predicts when a learning algorithm will perform best on a domain given features describing the problem (Aha, 1992). A recent special issue of the *Machine Learning* journal on "Bias Evaluation and Selection" introduced by Gordon and desJardins (1995) presents current research in this general area.

## Learning to Disambiguate Word Senses

Several recent research projects have taken a corpus-based approach to lexical disambiguation (Brown, Della-Pietra, Della-Pietra, & Mercer, 1991; Gale, Church, & Yarowsky, 1992b; Leacock et al., 1993b; Lehman, 1994). The goal is to learn to use surrounding context to determine the sense of an ambiguous word. Our tests are based on the corpus assembled by Leacock et al. (1993b). The task is to disambiguate the word "line" into one of six possible senses (text, formation, division, phone, cord, product) based on the words occurring in the current and previous sentence. The corpus was assembled from the 1987-89 *Wall Street Journal* and a 25 million word corpus from the American Printing House for the Blind. Sentences containing "line" were extracted and assigned a single sense from WordNet (Miller, 1991). There are a total of 4,149 examples in the full corpus unequally distributed across the six senses. Due to the use of the *Wall Street Journal*, the "product" sense is more than 5 times as common as any of the others. Previous studies have first sampled the data so that all senses were equally represented.

---

[1] Although multi-layer networks with sufficient hidden can represent arbitrary nonlinear functions, they will tend to learn a linear function if one exists that is consistent with the training data.

[2] This explanation was originally presented by Shavlik et al. (1991).

Leacock et al. (1993b), Leacock, Towell, and Voorhees (1993a) and Voorhees, Leacock, and Towell (1995) present results on a Bayesian method (Gale, Church, & Yarowsky, 1992a), a *content vector* method from information retrieval (Salton, Wong, & Yang, 1975), and a neural network trained using backpropagation (Rumelhart, Hinton, & Williams, 1986). The neural network architecture that performed at least as well as any other contained no hidden units, so was effectively equivalent to a perceptron. On the six-sense task trained on 1,200 examples and averaged over three random trials, they report the following generalization accuracies: Bayesian, 71%; content vectors, 72%; neural nets, 76%. None of these differences were statistically significant given the small number of trials.

In these studies, the data for the content-vector and neural-network methods was first reduced by ignoring case and reducing words to stems (e.g. *computer(s), computing, computation(al)*, etc. are all conflated to the feature *comput*) and removing a set of about 570 high-frequency *stopwords* (e.g. *the, by, you*, etc.). Similar preprocessing was performed for the current experiments, but we can not guarantee identical results. The result was a set of 2,094 examples equally distributed across the six senses where each example was described using 2,859 binary features each representing the presence or absence of a particular word stem in the current or immediately preceding sentence.

## Learning Algorithms Tested

The current experiments test a total of seven different learning algorithms with quite different biases. This section briefly describes each of these algorithms. Except for C4.5, which uses the C code provided by Quinlan (1993), all of these methods are implemented in Common Lisp and available on-line at http://www.cs.utexas.edu/users/ml/ml-progs.html. All systems were run on a Sun SPARCstation 5 with 40MB of main memory.

The simplest algorithms tested were a naive Bayesian classifier which assumes conditional independence of features and a $k$ nearest-neighbor classifier, which assigns a test example to the majority class of the 3 closest training examples (using Hamming distance to measure closeness) (Duda & Hart, 1973; Kulikowski & Weiss, 1991). Initial results indicated that $k$ nearest neighbor with $k=3$ resulted in slightly better performance than $k=1$. Naive Bayes is intended as a simple representative of statistical methods and nearest neighbor as a simple representative of instance-based (case-based, exemplar) methods (Cover & Hart, 1967; Aha, Kibler, & Albert, 1991).

Since the previous results of Leacock et al. (1993b) indicated that neural networks did not benefit from hidden units on the "line" disambiguation data, we employed a simple perceptron (Rosenblatt, 1962) as a representative connectionist method. The implementation learns a separate perceptron for recognizing each sense and assigns a test case to the sense indicated by the perceptron whose output most exceeds its threshold. In the current experiments, there was never a problem with convergence during training.

As a representative of decision-tree methods, we chose C4.5 (Quinlan, 1993), a system that is easily available and included in most recent experimental comparisons in machine learning. All parameters were left at their default values. We also tested C4.5-RULES, a variant of C4.5 in which decision trees are translated into rules and pruned; however, its performance was slightly inferior to the base C4.5 system on the "line" corpus; therefore, its results are not included.

Finally, we tested three simple logic-based induction algorithms that employ different representations of concepts: DNF, CNF, and decision lists. Most rule-based methods, e.g. Michalski (1983), induce a disjunctive set of conjunctive rules and therefore represent concepts in DNF. Some recent results have indicated that representing concepts in CNF (a conjunction of disjunctions) frequently performs somewhat better (Mooney, 1995). Some concepts are more compactly represented in CNF compared to DNF and vice versa. Therefore, both representations are included. Finally, *decision lists* (Rivest, 1987) are ordered lists of conjunctive rules, where rules are tested in order and the first one that matches an instance is used to classify it. A number of effective concept-learning systems have employed decision lists (Clark &

Niblett, 1989; Quinlan, 1993; Mooney & Califf, 1995) and they have already been successfully applied to lexical disambiguation (Yarowsky, 1994).

All of the logic-based methods are variations of the FOIL algorithm for induction of first-order function-free Horn clauses (Quinlan, 1990), appropriately simplified for the propositional case. They are called PFOIL-DNF, PFOIL-CNF, and PFOIL-DLIST. The algorithms are greedy covering (separate-and-conquer) methods that use an information-theoretic heuristic to guide a top-down search for a simple definition consistent with the training data. PFOIL-DNF (PFOIL-CNF) learns a separate DNF (CNF) description for each sense using the examples of that sense as positive instances and the examples of all other senses as negative instances. Mooney (1995) describes PFOIL-DNF and PFOIL-CNF in more detail and PFOIL-DLIST is based on the first-order decision-list learner described by Mooney and Califf (1995).

## Experiments

In order to evaluate the performance of these seven algorithms, direct multi-trial comparisons on identical training and test sets were run on the "line" corpus. Such head-to-head comparisons of methods are unfortunately relatively rare in the empirical natural-language literature, where papers generally report results of a single method on a single training set with, at best, indirect comparisons to other methods.

### Experimental Methodology

Learning curves were generated by splitting the preprocessed "line" corpus into 1,200 training examples and 894 test cases, training all methods on an increasingly larger subset of the training data and repeatedly testing them on the test set. Learning curves are fairly common in machine learning but not in corpus-based language research. We believe they are important since they reveal how algorithms perform with varying amounts of training data and how their performance improves with additional training. Results on a fixed-sized training set gives only one data point on the learning curve and leaves the possibility that differences between algorithms are hidden due to a ceiling effect, in which there are sufficient training examples for all methods to reach near Bayes-optimal performance.[3] Learning curves generally follow a power law where predictive accuracy climbs fairly rapidly and then levels off at an asymptotic level. A learning curve can reveal whether the performance of a system is approaching an asymptote or whether additional training data would likely result in significant improvement. Since gathering annotated training data is an expensive time-consuming process, it is important to understand the performance of methods given varying amounts of training data.

---

[3] Bayes-optimal performance is achieved by always picking the category with the maximum probability given all of its features. This requires actually knowing the conditional probability of each category given each of the exponentially large number of possible instance descriptions.

In addition to measuring generalization accuracy, we also collected data on the CPU time taken to train and test each method for each training-set size measured on the learning curve. This provides information on the computational resources required by each method, which may also be useful in deciding between them for particular applications. It also provides data on how the algorithm scales by providing information on how training time grows with training-set size.

Finally, all results are averaged over ten random selections of training and test sets. The performance of a system can vary a fair bit from trial to trial, and a difference in accuracy on a single training set may not indicate an overall performance advantage. Unfortunately, most results reported in empirical natural-language research present only a single or very small number of trials. Running multiple trials also allows for statistical testing of the significance of any resulting differences in average performance. We employ a simple two-tailed, paired t-test to compare the performance of two systems for a given training-set size, requiring significance at the 0.05 level. Even more sophisticated statistical analysis of the results is perhaps warranted.

### Experimental Results

The resulting learning curves are shown in Figure 1 and results on training and testing time are shown in Figures 2 and 3. Figure 3 presents the time required to classify the complete set of 894 test examples.

With respect to accuracy, naive Bayes and perceptron perform significantly better ($p \leq 0.05$) than all other methods for all training-set sizes. Naive Bayes and perceptron are not significantly different, except at 1,200 training examples where naive Bayes has a slight advantage. Note that the results for 1,200 training examples are comparable to those obtained by Leacock et al. (1993b) for similar methods. PFOIL-DLIST is always significantly better than PFOIL-DNF and PFOIL-CNF and significantly better than 3 Nearest Neighbor and C4.5 at 600 and 1,200 training examples. C4.5 and 3 Nearest Neighbor are always significantly better than PFOIL-DNF and PFOIL-CNF but not significantly different from each other. Finally, PFOIL-DNF is significantly better than PFOIL-CNF at 600 and 1,200 training examples.

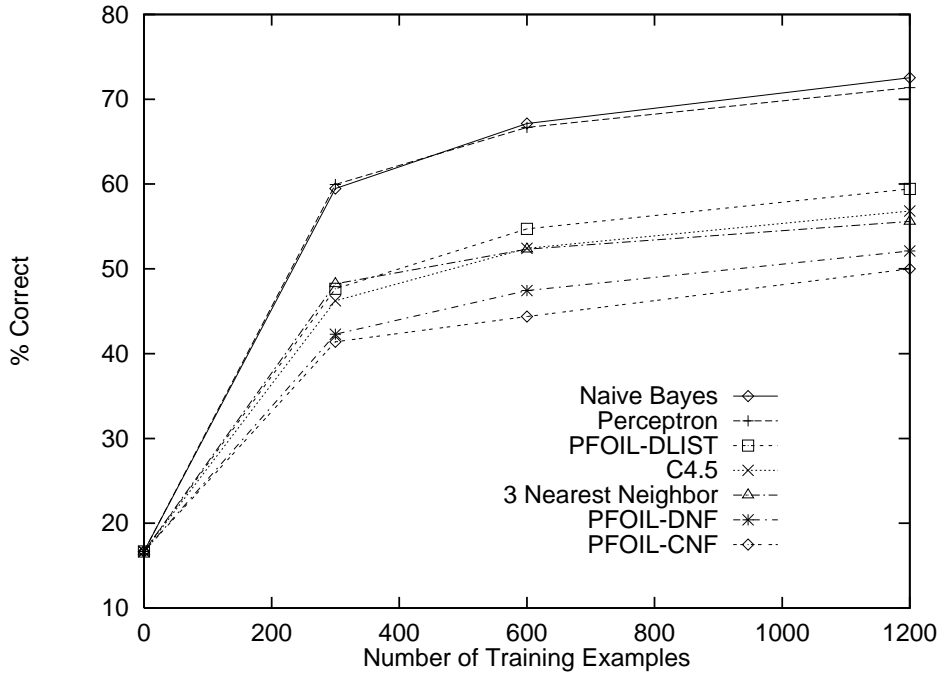

Figure 1: Accuracy at Disambiguating "Line"

With respect to training time, virtually all differences are significant. The logic-based induction methods are slowest, C4.5 and perceptron intermediate, and naive Bayes the fastest. Since it just stores examples, training time for Nearest Neighbor is always zero. In general, connectionist methods are much slower to train than alternative techniques (Shavlik et al., 1991); however, in this case a simple perceptron converges quite rapidly.

With respect to testing time, the symbolic induction methods are fastest and almost indistinguishable from zero in Figure 3 since they only need to test a small subset of the features. [4] All visible differences in the graph are significant. Naive Bayes is the slowest; both it and perceptron have the constant overhead of computing a weighted function over all of the almost 3,000 features. Nearest neighbor grows linearly with the number of training instances as expected; more sophisticated indexing methods can reduce this to logarithmic expected time (Friedman, Bentley, & Finkel, 1977).[5]

---

[4]C4.5 suffers a small constant overhead due to the C code having to read the test data in from a separate file.

[5]It should be noted that the implementation of nearest neighbor was optimized to handle sparse binary vectors by only including and comparing the features actually present in the examples. Without this optimization, testing would have been several orders of magnitude slower.

## Discussion of Results

Naive Bayes and perceptron are similar in that they both employ a weighted combination of all features. The decision-tree and logic-based approaches all attempt to find a combination of a relatively small set of features that accurately predict classification. After training on 1,200 examples, the symbolic structures learned for the line corpus are relatively large. Average sizes are 369 leaves for C4.5 decision trees, 742 literals for PFOIL-DLIST decision lists, 841 literals for PFOIL-DNF formulae, and 1197 literals for PFOIL-CNF formulae. However, many nodes or literals can test the same feature and the last two results include the total literal count for six separate DNF or CNF formulae (one for each sense). Therefore, each discrimination is clearly only testing a relatively small fraction of the 2,859 available features. Nearest neighbor bases its classifications on all features; however, it weights them all equally. Therefore, differential weighting is apparently necessary for high-performance on this problem. Alternative instance-based methods that weight features based on their predictive ability have also been developed (Aha et al., 1991). Therefore, our results indicate that lexical disambiguation is perhaps best performed using methods that combine weighted evidence from all of the features rather

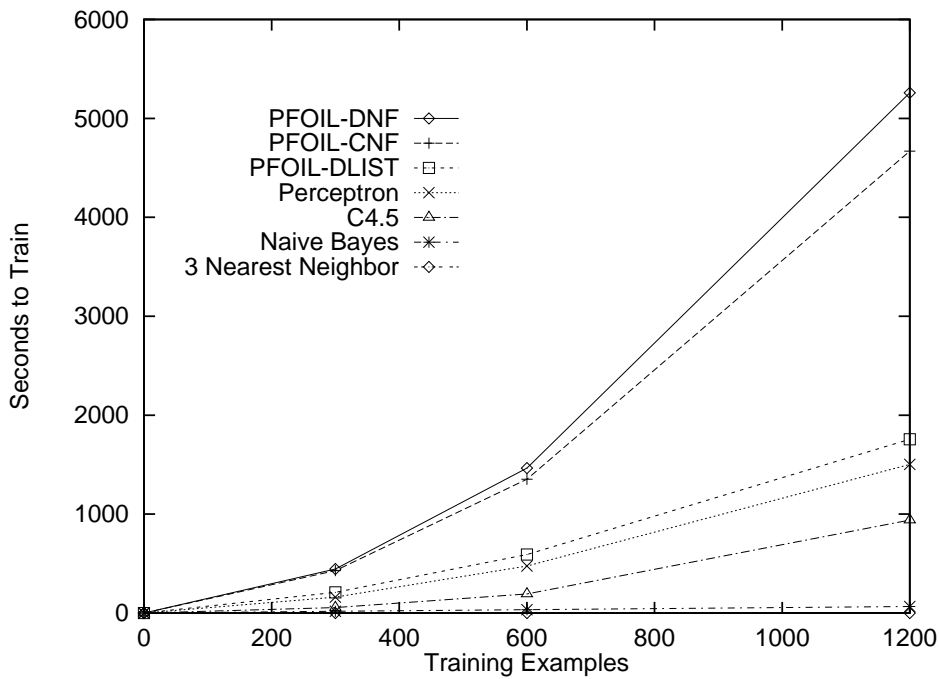

Figure 2: Training Time for "Line" Corpus

than making a decision by testing only a small subset of highly predictive features.

Among the other methods tested, decision lists seem to perform the best. The ordering of rules employed in a decision list in order to simplify the representation and perform conflict resolution apparently gives it an advantage over other symbolic methods on this task. In addition to the results reported by Yarowsky (1994) and Mooney and Califf (1995), it provides evidence for the utility of this representation for natural-language problems.

With respect to training time, the symbolic methods are significantly slower since they are searching for a simple declarative representation of the concept. Empirically, the time complexity for most methods are growing somewhat worse than linearly in the number of training examples. The worst in this regard are PFOIL-DNF and PFOIL-CNF which have a worst-case complexity of $O(n^2)$ (Mooney, 1995). However, all of the methods are able to process fairly large sets of data in reasonable time.

With respect to testing time, the symbolic methods perform the best since they only need to test a small number of features before making a decision. Therefore, in an application where response time is critical, learned rules or decision trees could provide rapid classification with only a modest decrease in accuracy. Not surprisingly, there is a trade-off between training time and testing time, the symbolic methods spend more effort during training compressing the representation of the learned concept resulting in a simpler description that is quicker to test.

## Future Research

The current results are for only one simple encoding of the lexical disambiguation problem into a feature vector representing an unordered set of word stems. This paper has focused on exploring the space of possible algorithms rather than the space of possible input representations. Alternative encodings which exploit positional information, syntactic word tags, syntactic parse trees, semantic information, etc. should be tested to determine the utility of more sophisticated representations. In particular, it would be interesting to see if the accuracy ranking of the seven algorithms is affected by a change in the representation.

Similar comparisons of a range of algorithms should also be performed on other natural language problems such as part-of-speech tagging (Church, 1988), prepositional phrase attachment (Hindle & Rooth, 1993), anaphora resolution (Anoe & Bennett, 1995), etc.. Since the requirements of individual tasks vary, different algorithms may be suitable for different sub-problems in natural language processing.

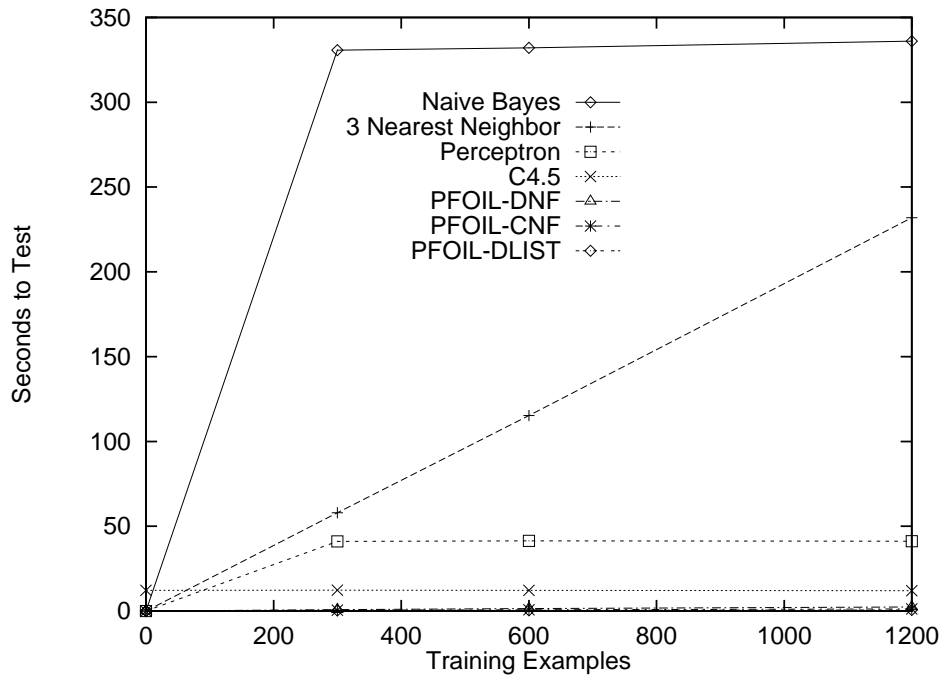

Figure 3: Testing Time for "Line" Corpus

## Conclusions

This paper has presented fairly comprehensive experiments comparing seven quite different empirical methods on learning to disambiguate words in context. Methods that employ a weighted combination of a large set of features, such as simple Bayesian and neural-network methods, were shown to perform better than alternative methods such as decision-tree, rule-based, and instance-based techniques on the problem of disambiguating the word "line" into one of six possible senses given the words that appear in the current and previous sentence as context. Although different learning algorithms can frequently perform quite similarly, they all have specific biases in their representation of concepts and therefore can illustrate both strengths and weaknesses in particular applications. Only rigorous experimental comparisons together with a qualitative analysis and explanation of their results can help determine the appropriate methods for particular problems in natural language processing.

## Acknowledgements

This research was partially supported by the National Science Foundation through grant IRI-9310819. I would also like to thank Goeff Towell for providing access to the "line" corpus.

## References

Aha, D. W. (1992). Generalizing from case studies: A case study. In *Proceedings of the Ninth International Conference on Machine Learning*, pp. 1–10 Aberdeen, Scotland.

Aha, D. W., Kibler, D., & Albert, M. K. (1991). Instance-based learning algorithms. *Machine Learning*, 6(1), 37–66.

Anoe, C., & Bennett, S. W. (1995). Evaluating automated and manual acquisition of anaphora resolution strategies. In *Proceedings of the 33rd Annual Meeting of the Association for Computational Linguistics*, pp. 122–129 Cambridge, MA.

Atlas, L., Cole, R., Conner, J., El-Sharkawi, M., Marks, R., Muthusamy, Y., & Bernard, E. (1990). Performance comparisons between backpropagation networks and classification trees on three real-world applications. In Touretzky, D. S. (Ed.), *Advances in Neural Information Processing Systems 2*. Morgan Kaufmann, San Mateo, CA.

Baffes, P., & Mooney, R. (1993). Symbolic revision of theories with M-of-N rules. In *Proceedings of the Thirteenth International Joint Conference on Artificial Intelligence*, pp. 1135–1140 Chambery, France.

Blumer, A., Ehrenfeucht, A., Haussler, D., & Warmuth, M. (1987). Occam's razor. *Information Processing Letters*, 24, 377–380.

Brill, E. (1993). Automatic grammar induction and parsing free text: A transformation-based approach. In *Proceedings of the 31st Annual Meeting of the Association for Computational Linguistics*, pp. 259–265 Columbus, Ohio.

Brown, P., Della-Pietra, S., Della-Pietra, V., & Mercer, R. (1991). Word sense disambiguation using statistical methods. In *Proceedings of the 29th Annual Meeting of the Association for Computational Linguistics*, pp. 264–270.

Charniak, E. (1993). *Statistical Language Learning*. MIT Press.

Church, K. (1988). A stochastic parts program and noun phrase parser for unrestricted text. In *Proceedings of the Second Conference on Applied Natural Language Processing*. Association for Computational Linguistics.

Clark, P., & Niblett, T. (1989). The CN2 induction algorithm. *Machine Learning, 3*, 261–284.

Cover, T. M., & Hart, P. E. (1967). Nearest neighbor pattern classification. *IEEE Transactions on Information Theory, 13*, 21–27.

Dietterich, T. G., Hild, H., & Bakiri, G. (1990). A comparative study of ID3 and backpropagation for English text-to-speech mapping. In *Proceedings of the Seventh International Conference on Machine Learning*, pp. 24–31 Austin, TX.

Duda, R. O., & Hart, P. E. (1973). *Pattern Classification and Scene Analysis*. Wiley, New York.

Fisher, D. H., & McKusick, K. B. (1989). An empirical comparison of ID3 and backpropagation. In *Proceedings of the Eleventh International Joint Conference on Artificial Intelligence*, pp. 788–793 Detroit, MI.

Friedman, J., Bentley, J., & Finkel, R. (1977). An algorithm for finding best matches in logarithmic expected time. *ACM Transactions on Mathematical Software, 3*(3), 209–226.

Gale, W., Church, K., & Yarowsky, D. (1992a). A method for disambiguating word senses in a large corpus. *Computers and the Humanities, 26*, 415–439.

Gale, W., Church, K. W., & Yarowsky, D. (1992b). Estimating upper and lower bounds on the performance of word-sense disambiguation programs. In *Proceedings of the 30th Annual Meeting of the Association for Computational Linguistics*, pp. 249–256 Newark, Delaware.

Gordon, D. F., & desJardins, M. (1995). Evaluation and selection of biases in machine learning. *Machine Learning, 20*(1/2), 5–22.

Hindle, D., & Rooth, M. (1993). Structural ambiguity and lexical relations. *Computational Linguistics, 19*(1), 103–120.

Holte, R. C. (1993). Very simple classification rules perform well on most commonly used datasets. *Machine Learning, 11*(1), 63–90.

Hume, D. (1748). *An Inquiry Concerning Human Understanding* Reprinted 1955. Liberal Arts Press, New York.

Kulikowski, C. A., & Weiss, S. M. (1991). *Computer Systems That Learn - Classification and Prediction Methods from Statistics, Neural Nets, Machine Learning, and Expert Systems*. Morgan Kaufmann, San Mateo, CA.

Langley, P. (1996). *Elements of Machine Learning*. Morgan Kaufmann, San Francisco, CA.

Leacock, C., Towell, G., & Voorhees, E. (1993a). Corpus-based statistical sense resolution. In *Proceedings of the ARPA Workshop on Human Language Technology*.

Leacock, C., Towell, G., & Voorhees, E. (1993b). Towards building contextual representations of word senses using statistical models. In *Proceedings of the SIGLEX Workshop: Acquisition of Lexical Knowledge from Text*, pp. 10–20. Association for Computational Linguistics.

Lehman, J. F. (1994). Toward the essential nature of satistical knowledge in sense resolution. In *Proceedings of the Twelfth National Conference on Artificial Intelligence*, pp. 734–741 Seattle, WA.

Ling, C. X. (1994). Learning the past tense of English verbs: The symbolic pattern associator vs. connectionist models. *Journal of Artificial Intelligence Research, 1*, 209–229.

Ling, C. X., & Marinov, M. (1993). Answering the connectionist challenge: A symbolic model of learning the past tense of English verbs. *Cognition, 49*(3), 235–290.

Marcus, M., Santorini, B., & Marcinkiewicz, M. (1993). Building a large annotated corpus of English: The Penn treebank. *Computational Linguistics, 19*(2), 313–330.

Merz, C., Murphy, P. M., & Aha, D. W. (1996). Repository of machine learning databases http://www.ics.uci.edu/~mlearn/mlrepository.html. Department of Information and Computer Science, University of California, Irvine, CA.

Michalski, R. S. (1983). A theory and methodology of inductive learning. In Michalski,

R. S., Carbonell, J. G., & Mitchell, T. M. (Eds.), *Machine Learning: An Artificial Intelligence Approach*, pp. 83–134. Tioga.

Miller, G. (1991). WordNet: An on-line lexical database. *International Journal of Lexicography*, *3*(4).

Mitchell, T. (1980). The need for biases in learning generalizations. Tech. rep. CBM-TR-117, Rutgers University. Reprinted in *Readings in Machine Learning*, J. W. Shavlik and T. G. Dietterich (eds.), Morgan Kaufman, San Mateo, CA, 1990.

Mooney, R. J. (1995). Encouraging experimental results on learning CNF. *Machine Learning*, *19*(1), 79–92.

Mooney, R. J., & Califf, M. E. (1995). Induction of first-order decision lists: Results on learning the past tense of English verbs. *Journal of Artificial Intelligence Research*, *3*, 1–24.

Mooney, R. J., Shavlik, J. W., Towell, G., & Gove, A. (1989). An experimental comparison of symbolic and connectionist learning algorithms. In *Proceedings of the Eleventh International Joint Conference on Artificial Intelligence*, pp. 775–780 Detroit, MI. Reprinted in *Readings in Machine Learning*, J. W. Shavlik and T. G. Dietterich (eds.), Morgan Kaufman, San Mateo, CA, 1990.

Murphy, P. M., & Pazzani, M. J. (1991). ID2-of-3: Constructive induction of M-of-N concepts for discriminators in decision trees. In *Proceedings of the Eighth International Workshop on Machine Learning*, pp. 183–187 Evanston, IL.

Pagallo, G., & Haussler, D. (1990). Boolean feature discovery in empirical learning. *Machine Learning*, *5*, 71–100.

Periera, F., & Shabes, Y. (1992). Inside-outside reestimation from partially bracketed corpora. In *Proceedings of the 30th Annual Meeting of the Association for Computational Linguistics*, pp. 128–135 Newark, Delaware.

Quinlan, J. R. (1993). *C4.5: Programs for Machine Learning*. Morgan Kaufmann, San Mateo,CA.

Quinlan, J. (1990). Learning logical definitions from relations. *Machine Learning*, *5*(3), 239–266.

Rao, R. B., Gordon, D., & Spears, W. (1995). For every generalization action is there really an equal an opposite reaction? Analysis of the conservation law for generalization performance. In *Proceedings of the Twelfth International Conference on Machine Learning*, pp. 471–479 San Francisco, CA. Morgan Kaufman.

Reilly, R. G., & Sharkey, N. E. (Eds.). (1992). *Connectionist Approaches to Natural Language Processing*. Lawrence Erlbaum and Associates, Hilldale, NJ.

Rivest, R. L. . (1987). Learning decision lists. *Machine Learning*, *2*(3), 229–246.

Rosenblatt, F. (1962). *Principles of Neurodynamics*. Spartan, New York.

Rumelhart, D. E., Hinton, G. E., & Williams, J. R. (1986). Learning internal representations by error propagation. In Rumelhart, D. E., & McClelland, J. L. (Eds.), *Parallel Distributed Processing, Vol. I*, pp. 318–362. MIT Press, Cambridge, MA.

Salton, G., Wong, A., & Yang, C. S. (1975). A vector space model for automatic indexing. *Communications of the Association for Computing Machinery*, *18*(11), 613–620.

Schaffer, C. (1993). Selecting a classification method by cross-validation. *Machine Learning*, *13*(1), 135–143.

Schaffer, C. (1994). A conservation law for generalization performance. In *Proceedings of the Eleventh International Conference on Machine Learning*, pp. 259–265 San Francisco, CA. Morgan Kaufman.

Shavlik, J. W., & Dietterich, T. G. (Eds.). (1990). *Readings in Machine Learning*. Morgan Kaufmann, San Mateo,CA.

Shavlik, J. W., Mooney, R. J., & Towell, G. G. (1991). Symbolic and neural learning algorithms: An experimental comparison. *Machine Learning*, *6*, 111–143. Reprinted in *Readings in Knowledge Acquisition and Learning*, B. G. Buchanan and D. C. Wilkins (eds.), Morgan Kaufman, San Mateo, CA, 1993.

Towell, G. G., Shavlik, J. W., & Noordewier, M. O. (1990). Refinement of approximate domain theories by knowledge-based artificial neural networks. In *Proceedings of the Eighth National Conference on Artificial Intelligence*, pp. 861–866 Boston, MA.

Voorhees, E., Leacock, C., & Towell, G. (1995). Learning context to disambiguate word senses. In Petsche, T., Hanson, S., & Shavlik, J. (Eds.), *Computational Learning Theory and Natural Learning Systems, Vol. 3*, pp. 279–305. MIT Press, Cambridge, MA.

Weiss, S. M., & Kapouleas, I. (1989). An empirical comparison of pattern recognition, neural nets, and machine learning classification methods. In *Proceedings of the Eleventh International Joint Conference on Artificial Intelligence*, pp. 781–787 Detroit, MI.

Wermter, S., Riloff, E., & Scheler, G. (Eds.). (1996). *Symbolic, Connectionist, and Statistical Approaches to Learning for Natural Language Processing*. Springer Verlag, Berlin. in press.

Wolpert, D. H. (1992). On the connection between in-sample testing and generalization error. *Complex Systems*, *6*, 47–94.

Yarowsky, D. (1994). Decision lists for lexical ambiguity resolution: Application to accent restoration in Spanish and French. In *Proceedings of the 32nd Annual Meeting of the Association for Computational Linguistics*, pp. 88–95 Las Cruces, NM.

Zelle, J. M., & Mooney, R. J. (1994). Inducing deterministic Prolog parsers from treebanks: A machine learning approach. In *Proceedings of the Twelfth National Conference on Artificial Intelligence*, pp. 748–753 Seattle, WA.